\documentclass[10pt, a4paper]{article}
\usepackage{lrec}

\usepackage{bm,amsmath}
\usepackage{mathtools}
\usepackage{anyfontsize}
\usepackage{multirow}

\usepackage{graphicx}
\usepackage{tabularx}
\usepackage{soul}
\usepackage{arydshln}
\usepackage{float}

\usepackage{epstopdf}
\usepackage{inputenc}

\usepackage{hyperref}
\usepackage{xstring}

\usepackage{flushend}

\def\Underline{\setbox0\hbox\bgroup\let\\\endUnderline}
\def\endUnderline{\vphantom{y}\egroup\smash{\underline{\box0}}\\}
\def\|{\verb|}

\title{Speech Corpus of Ainu Folklore and End-to-end Speech Recognition\\for Ainu Language}

\name{Kohei Matsuura, Sei Ueno, Masato Mimura, Shinsuke Sakai, Tatsuya Kawahara}

\address{Graduate School of Informatics, Kyoto University\\
         Sakyo-ku,  Kyoto 606-8501, Japan\\
         \{matsuura, ueno, mimura, sakai, kawahara\}@sap.ist.i.kyoto-u.ac.jp}

\abstract{
Ainu is an unwritten language that has been spoken by Ainu people who are one of the ethnic groups in Japan.
It is recognized as critically endangered by UNESCO and archiving and documentation of its language heritage is of paramount importance.
Although a considerable amount of voice recordings of Ainu folklore has been produced and accumulated to save their culture, only a quite limited parts of them are transcribed so far. 
Thus, we started a project of automatic speech recognition (ASR) for the Ainu language in order to contribute to the development of annotated language archives.
In this paper, we report speech corpus development and the structure and performance of end-to-end ASR for Ainu.
We investigated four modeling units (phone, syllable, word piece, and word) and found that the syllable-based model performed best in terms of both word and phone recognition accuracy, which were about 60\% and over 85\% respectively in speaker-open condition.
Furthermore, word and phone accuracy of 80\% and 90\% has been achieved in a speaker-closed setting.
We also found out that a multilingual ASR training with additional speech corpora of English and Japanese further improves the speaker-open test accuracy.
}

\begin{document}

\maketitleabstract

\section{Introduction}
Automatic speech recognition (ASR) technology has been made a dramatic progress and is currently brought to a pratical levels of performance assisted by large speech corpora and the introduction of deep learning techniques.
However, this is not the case for low-resource languages which do not have large corpora like English and Japanese have. There are about 5,000 languages in the world over half of which are faced with the danger of extinction. Therefore, constructing ASR systems for these endangered languages is an important issue.

The Ainu are an indigenous people of northern Japan and Sakhakin in Russia, 
but their language has been fading away ever since the Meiji Restoration and Modernization. 
On the other hand, active efforts to preserve their culture have been initiated by the Government of Japan, and exceptionally large oral recordings have been made. Nevertheless, a majority of the recordings have not been transcribed and utilized effectively. Since transcribing them requires expertise in the Ainu language, not so many people are able to work on this task. Hence, there is a strong demand for an ASR system for the Ainu language. We started a project of Ainu ASR and this article is the first report of this project.

We have built an Ainu speech corpus based on data provided by the Ainu Museum\footnote{http://ainugo.ainu-museum.or.jp/} and the Nibutani Ainu Culture Museum\footnote{http://www.town.biratori.hokkaido.jp/biratori/nibutani/}. The oral recordings in this data consist of folklore and folk songs, and we chose the former to construct the ASR model. The end-to-end method of speech recognition has been proposed recently and has achieved performance comparable to that of the conventional DNN-HMM hybrid modeling \cite{sotaGoogle,Povey2018,han2019stateoftheart}.  End-to-end systems do not have a complex hierarchical structure and do not require expertise in target languages such as their phonology and morphology. In this study we adopt the attention mechanism \cite{6af3452a28a04980b2b8f5eb48730d36,AttnProto}  and combine it with Connectionist Temporal Classification  (CTC) \cite{GravesCTC,Graves2014TowardsES}. In this work, we investigate the modeling unit and utilization of corpora of other languages. 

\begin{centering}
\begin{table*}[t]
\centering
\caption{Speaker-wise details of the corpus} \vspace{5pt}
\scalebox{1.01}[1.01]{
\begingroup
\renewcommand{\arraystretch}{1.2} 
\begin{tabular}{l|cccccccc|c} \hline \hline
\multicolumn{1}{c|}{Speaker ID} & KM  & UT   & KT& HS   & NN   & KS& HY& KK   & total\\ \hline 
duration (h:m:s)  & 19:40:58&7:14:53&3:13:37&2:05:39&1:44:32&1:43:29&1:36:35&1:34:55&38:54:38\\ 
duration (\%) & 50.6 & 18.6 & 8.3 & 5.4 & 4.5 & 4.4 & 4.1 & 4.1 & 100.0\\
\# episodes  & 29& 26 & 20  & 8& 8 & 11   & 8  & 7& 114\\
\# IPUs & 9170   & 3610 & 2273 & 2089 & 2273 & 1302 & 1220  & 1109  & 22345\\ \hline
\end{tabular}
\endgroup}
\end{table*}
\end{centering}

\section{Overview of the Ainu Language}
This section briefly overviews the background of the data collection, the Ainu language, and its writing system. After that, we describe how Ainu recordings are classified and review previous works dealing with the Ainu language.

\subsection{Background}
The Ainu people had total population of about 20,000 in the mid-19th century \cite{hardacre1997new} and they used to live widely distributed in the area that includes Hokkaido, Sakhalin, and the Kuril Islands.
The number of native speakers, however, rapidly decreased through the assimilation policy after late 19th century. At present, there are only less than 10 native speakers, and UNESCO listed their language as critically endangered in 2009 \cite{unesco}. 
In response to this situation, Ainu folklore and songs have been actively recorded since the late 20th century in efforts initiated by the Government of Japan. 
For example, the Ainu Museum started audio recording of Ainu folklore in 1976 with the cooperation of a few Ainu elders which resulted in the collection of speech data with the total duration of roughly 700 hours.
This kind of data should be a key to the understanding of Ainu culture, but most of it is not transcribed and fully studied yet.

\subsection{The Ainu Language and its Writing System}
The Ainu language is an agglutinative language and has some similarities to Japanese. However, its genealogical relationship with other languages has not been clearly understood yet.
Among its features such as closed syllables and personal verbal affixes, one important feature is that there are many compound words. For example, a word \textit{atuykorkamuy} (means ``a sea turtle'') can be disassembled into \textit{atuy} (``the sea''), \textit{kor} (``to have''), and \textit{kamuy} (``god'').

Although the Ainu people did not traditionally have a writing system, the Ainu language is currently written following the examples in a reference book ``Akor itak'' \cite{akor}. With this writing system, it is transcribed with sixteen Roman letters \{a, c, e, h, i, k, m, n, o, p, r, s, t, u, w, y\}. Since each of these letters correspond to a unique pronunciation, we call them ``phones'' for convenience. 
In addition, the symbol \{=\} is used for connecting a verb and a personal affix and \{ ' \} is used to represent the pharyngeal stop.
For the purpose of transcribing recordings, consonant symbols \{b, d, g, z\} are additionally used to transcribe Japanese sounds the speakers utter. The symbols \{ \_ , \_\_ \} are used to transcribe drops and liaisons of phones. An example is shown below.

\begin{table}[h]
\centering
\begingroup 
\renewcommand{\arraystretch}{1.5} 
\begin{tabular}{c|ccccc} 
\it{original}&\multicolumn{5}{c}{mos=an \_\_hine inkar'=an}\\ 
\it{translation}&\multicolumn{5}{c}{I wake up and look}\\ \hline
\multirow{2}{*}{\it{structure}}&mos & =an  &hine & inkar & =an  \\ 
 & \bf{wake up}& \bf{1sg}& \bf{and}& \bf{look}& \bf{1sg}\\
\end{tabular}
\endgroup
\end{table}

\begin{centering}
\begin{table}[t]
\caption{Text excerpted from the prose tale `\textit{The Boy Who Became Porosir God}' spoken by KM.\vspace{5pt}}
\scalebox{0.88}[0.88]{
\begingroup
\renewcommand{\arraystretch}{1.4} 
\begin{tabular}{|l|l|} \hline
\multicolumn{1}{|c|}{\it{original}} & \multicolumn{1}{c|}{\it{English translation}}\\ \hline
Samormosir mosir & In neighboring country\\
noski ta & at the middle (of it),\\
a=kor hapo i=resu hine & being raised by my mother,\\
oka=an pe ne \_hike & I was leading my life.\\
kunne hene tokap \_hene & Night and day, all day long,\\
yam patek i=pareoyki & I was fed with chestnut\\
yam patek a=e kusu & and all I ate was chestnut,\\
somo hetuku=an pe ne kunak & so, that I would not grow up\\
a=ramu a korka & was my thought. \\ \hline
\end{tabular}
\endgroup

}\centering
\end{table}
\end{centering}

\subsection{Types of Ainu Recordings}
The Ainu oral traditions are classified into three types: ``\textit{yukar}'' (heroic epics), ``\textit{kamuy yukar}'' (mythic epics), and ``\textit{uwepeker}'' (prose tales). \textit{Yukar} and \textit{kamuy yukar} are recited in the rhythm while \textit{uwepeker} is not. In this study we focus on the the prose tales as the first step.

\subsection{Previous Work}
There have so far been a few studies dealing with the Ainu language. 
\newcite{ainulrec} built a dependency tree bank in the scheme of Universal Dependencies\footnote{https://universaldependencies.org/}. \newcite{postag} developed tools for part-of-speech (POS) tagging and word segmentation. Ainu speech recognition was tried by \newcite{ainutrans} with 2.5 hours of Ainu folklore data even though the Ainu language was not their main target. Their phone error rare was about 40\% which is not an accuracy level for practical use yet.

It appears that there has not been a substantial Ainu speech recognition study yet that utilizes corpora of a reasonable size.
Therefore, our first step was to build a speech corpus for ASR based on the data sets provided by the Ainu Museum and the Nibutani Ainu Culture Museum.

\section{Ainu Speech Corpus}
In this section we explain the content of the data sets and how we modified it for our ASR corpus.

\subsection{Numbers of Speakers and Episodes}
The corpus we have prepared for ASR in this study is composed of text and speech. Table 1 shows the number of episodes and the total speech duration for each speaker. Among the total of eight speakers, the data of the speakers KM and UT is from the Ainu Museum, and the rest is from Nibutani Ainu Culture Museum. 
All speakers are female.
The length of the recording for a speaker varies depending on the circumstances at the recording times.
A sample text and its English translation are shown in Table 2. 

\subsection{Data Annotation}
For efficient training of ASR model, we have made some modifications to the provided data. First, from the transcripts explained in Section 2.1, the symbols \{\_ , \_\_ , '\} have been removed as seen in the example below. 

\begin{table}[H]
\centering
\begingroup 
\renewcommand{\arraystretch}{1.2} 
\begin{tabular}{c|l} 
\it{original}   & uymam'=an wa isam=an \_\_hi okake ta \\ 
\it{modified} & uymam=an wa isam=an hi okake ta\\ 
\end{tabular}
\endgroup
\end{table}

Though the equal symbol (`=') does not represent a sound, we keep it because it is used in almost all of the Ainu documents and provides grammatical information.


To train an ASR system, the speech data needs to be segmented into a set of manageable chunks. For the ease of automatic processing, we chose to segment speech into \textit{inter-pausal units} (IPUs) \cite{IPU}which is a stretch of speech bounded by pauses.
The number of IPUs for each speaker is shown in Table 1.

\section{End-to-end Speech Recognition}
In this section, the two approaches to end-to-end speech recognition that we adopt in this work are summarized. 
Then, we introduce four modeling units we explained, i.e., phone, syllable, word piece, and word.
We also discuss multilingual training that we adopt for tackling the low resource problem.

\subsection{End-to-end Modeling}
End-to-end models have an architecture much simpler than that of conventional DNN-HMM hybrid models.
Since they predict character or word symbols directly from acoustic features, pronunciation dictionaries and language modeling are not required explicitly. 
In this paper, we utilize two kinds of end-to-end models, namely, Connectionist Temporal Classification (CTC) and the attention-based encoder-decoder model.

CTC augments the output symbol set with the ``blank'' symbol `$\phi$'.
It outputs symbols by contracting frame-wise outputs from recurrent neural networks (RNNs).
This is done by first collapsed repeating symbols and then removing all blank symbols as in the following example:

\begin{table}[H]
\centering
\scalebox{1.15}[1.15]{
\begin{tabular}{c}
aab$\phi$bbccc $\rightarrow$ abbc \vspace{-3pt}
\end{tabular}} 
\end{table}

\begin{figure}[t]
\centering
\includegraphics[width=1.0\columnwidth]{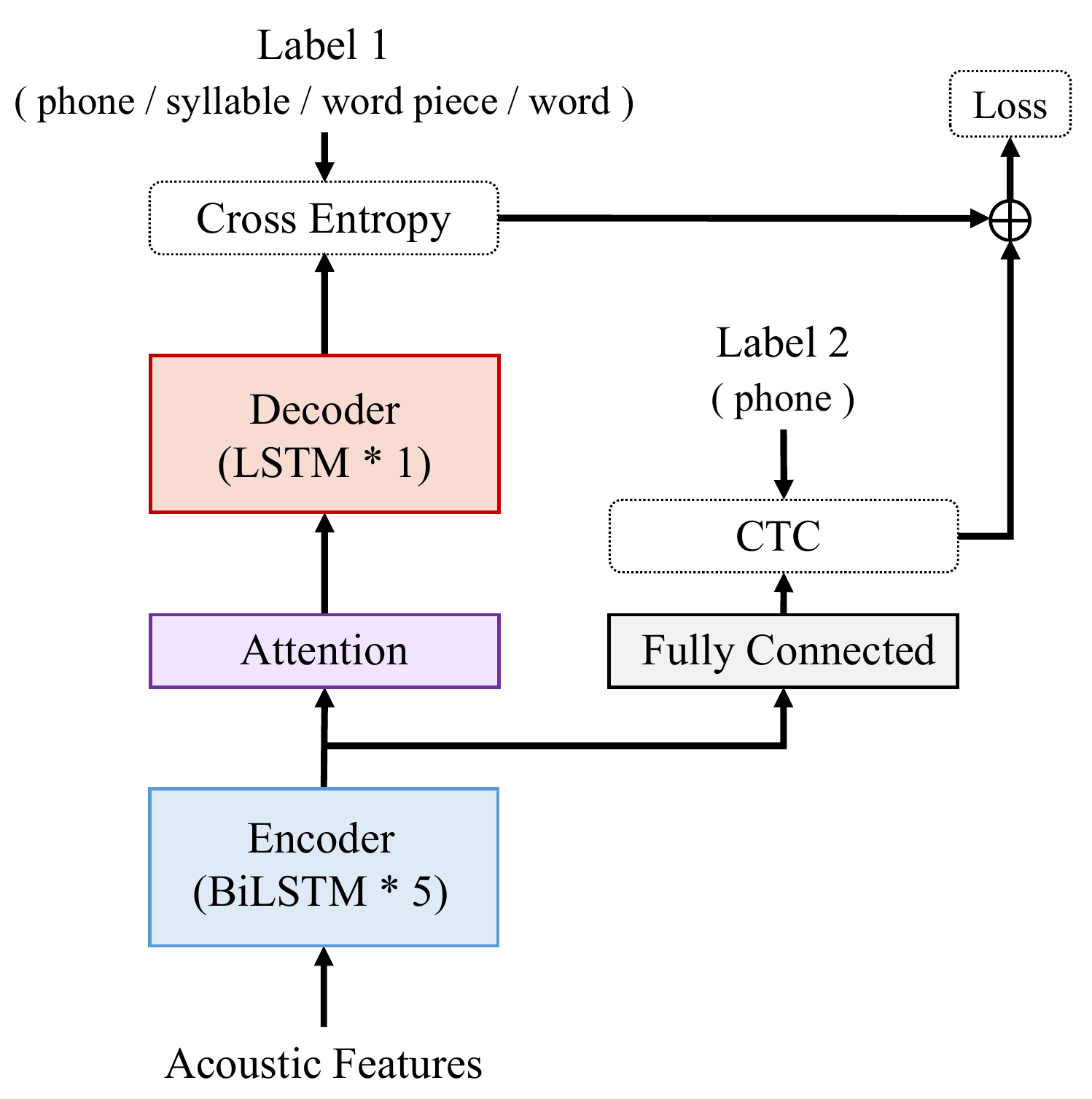}
\caption{The attention model with CTC auxiliary task.}
\label{fig:ctclas}
\end{figure}

The probability of an output sequence $\bm{L}$ for an input acoustic feature sequence $\bm{X}$, where $|\bm{L}| < |\bm{X}|$, is defined as follows.
\begin{align}
p(\bm{L}|\bm{X}) = \sum _ {\substack{\bm{\mathit{\Pi}} \in \mathcal{B}^{-1}(\bm{L})\\ |\bm{\mathit{\Pi}}| = |\bm{X}|}} p(\bm{\mathit{\Pi}} | \bm{X})
\end{align}

$\mathcal{B}$ is a function to contract the outputs of RNNs, so $\mathcal{B}^{-1}(\bm{L})$ means the set of symbol sequences which is reduced to $\bm{L}$. 
The model is trained to maximize (1).

The attention-based encoder-decoder model is another method for mapping between two sequences with different lengths. It has two RNNs called the ``encoder'' and the ``decoder''. 
In naive encoder-decoder model, the encoder converts the input sequence into a single context vector which is the last hidden state of the encoder RNN
from which the decoder infers output symbols. 
In an attention-based model, 
the context vector $\mathbf{c}_l$ at $l$-th decoding step is the sum of the product of all encoder outputs $\boldsymbol{h}_1, ... , \boldsymbol{h}_\mathrm{T}$ and the $l$-th attention weight $\alpha_{1,l}, ... , \alpha_{\mathrm{T},l}$ as shown in (2). Here, $\mathrm{T}$ is the length of the encoder output. 
\begin{eqnarray}
\boldsymbol{c}_l = \sum^\mathrm{T}_{t=1} \alpha_{t, l} \boldsymbol{h}_t
\end{eqnarray}
The attention weights $\boldsymbol{\alpha}_{1,l}, ... , \boldsymbol{\alpha}_{\mathrm{T},l}$ indicates the relative importances of the encoder output frames for the $l$-th decoding step and the model parameters to generate these weights are determined in an end-to-end training.

In our model, the attention-based model and the CTC share the encoder and are optimized simultaneously as shown in Figure 1.\cite{DBLP:journals/corr/KimHW16} Long Short-Term Memory (LSTM) \cite{Hochreiter:1997:LSM:1246443.1246450} is used for RNNs in the encoder and the decoder.

\subsection{Modeling Units}
In the conventional DNN-HMM hybrid modeling, the acoustic model outputs probabilities triphone states from each acoustic feature which is converted into the most likely word sequence.
An end-to-end model, on the other hand, has some degree of freedom in the modeling unit other than phones, and there are some studies that use characters or words as a unit \cite{LAS,Adv}. 
A word unit based end-to-end model can take long context into consideration at the inference time, but it has the data sparsity problem due to its large vocabulary size. Though phone unit based model does not have such a problem, it cannot grasp so long context. It depends on the size of available corpora to decide which to adopt.
In addition to these both models, a word piece unit, which is defined by automatically dividing a word into frequent parts, has been proposed \cite{WP,WP2}, and its vocabulary size can be determined almost freely.

In this paper, we investigate the modeling unit for the end-to-end Ainu speech recognition since the optimal unit for this size of corpus is not obvious. \cite{DBLP:journals/corr/abs-1902-01955}  It is presupposed that all units can be converted into word units automatically. The candidates are phone, syllable, word piece (WP), and word. Examples of them are shown in Table 3 and the details of each unit are described below. 

\subsubsection{Phone}
As mentioned in Section 2.1, we regard the Roman letters as phones. `=' and the special symbol `$\langle$wb$\rangle$', which means a word boundary, are added to make it possible to convert the output into a sequence of words like the `original' in Table 3.

\subsubsection{Syllable}
A syllable of the Ainu language takes the form of either V, CV, VC, or CVC, where `C' and `V' mean consonant and vowel, respectively. The phones \{a, e, i, o, u\} are vowels and the rest of the Roman letters in Section 2.2 are consonants. In this work, every word is divided into syllables by the following procedure.
\begin{enumerate}
\item{A word with a single letter is unchanged.}
\item{Two consecutive Cs and Vs are given a syllable boundary between them.\vspace{10pt}\\~~~~~~~~~~~~~{\fontsize{9}{11}\selectfont \sf{R$^*$\{CC, VV\}R$^*$$\rightarrow$~R$^*$\{C-C, V-V\}R$^*$}\\~~~~~~~~~~~~~~~~~~~~~~~~~~~~~~(\textsf{R} $\coloneqq$ \{\textsf{C}, \textsf{V}\})}}
\item{Put a syllable boundary after the segment-initial V if it is following by at least two phones.\vspace{10pt}\\~~~~~~~~~~~~~~~~~~~~~~~~~~~~{\fontsize{9}{11}\selectfont \sf{VCR$^+$}$\rightarrow$~\sf{V-CR$^+$}}}
\item{Put a syllable boundary after CV repeatedly from left to right until only CV or CVC is left.\vspace{10pt}\\~~~~~~~~~~{\fontsize{9}{11}\selectfont \sf{(CV)$^*$\{CV, CVC\}}~$\rightarrow$~\sf{(CV-)$^*$\{CV, CVC\}}}}
\end{enumerate}


\begin{table}[t]
\begin{center}
\centering
\caption{Examples of four modeling units.}\vspace{5pt}
\begingroup
\renewcommand{\arraystretch}{1.2} 
\begin{tabular}{c|l}
\hline \hline
original & a=saha i=kokopan wa \\ \hline
phone & a = s a h a $\langle$wb$\rangle$ i = k o k o p a n $\langle$wb$\rangle$ w a \\ 
syllable & a = sa ha  $\langle$wb$\rangle$ i = ko pan  $\langle$wb$\rangle$ wa \\ 
WP & $\langle$wb$\rangle$a = saha $\langle$wb$\rangle$i = ko p an $\langle$wb$\rangle$wa　\\ 
word & a = saha i = $\langle$unk$\rangle$ wa \\ \hline
translation & my elder sister told me not to do so  \\ \hline 
\end{tabular}
\endgroup
\end{center}
\vspace{-5pt}
\end{table}

In addition, `=' and `$\langle$wb$\rangle$' are added as explained in Section 4.2.1. through the model training process.

This procedure does not always generate a morphologically relevant syllable segmentation. For example, a word \textit{isermakus} (meaning ``(for a god) to protect from behind'') is divided as \textit{i-ser-ma-kus}, but the right syllabification is \textit{i-ser-mak-us}.

\subsubsection{Word Piece}
The byte pair encoding (BPE) \cite{DBLP:journals/corr/SennrichHB15} and the unigram language modeling \cite{kudo} are alternative methods for dividing a word into word pieces. The former repeatedly replaces the most common character pair with a new single symbol until the vocabulary becomes the intended size. The latter decides the segmentation to maximize the likelihood of occurrence of the sequence. We adopt the latter and use the open-source software SentencePiece\footnote{https://github.com/google/sentencepiece} \cite{sent}. 
With this tool, `$\langle$wb$\rangle$' and other units are often merged to constitute a single piece as seen in Table 3.

\subsubsection{Word}
The original text can be segmented into words separated by spaces. To make the vocabulary smaller for the ease of training, `=' is treated as a word and infrequent words are replaced with a special label `$\langle$unk$\rangle$'. As seen in Table 3, `\textit{a=saha}' is dealt with as three words (`\textit{a}', `\textit{=}', `\textit{saha}') and the word `\textit{kokopan}' is replaced with `$\langle$unk$\rangle$'.

\subsection{Multilingual Training}
When an enough amount of data is not available for the target languages,
the ASR model training  can be enhanced by taking advantage of data from other languages \cite{toshniwal2018multilingual,DBLP:journals/corr/abs-1810-03459}. 
There are some similarities between Ainu and Japanese language \cite{tamura-en}. For instance, both have almost the same set of vowels and do not have consonant clusters (like `\textit{str}' of `strike' in English). 
Hence, the multilingual training with a Japanese corpus is expected to be effective. In addition, an English corpus is used for the purpose of comparison. The corpora used are the JNAS corpus \cite{jnas} (in Japanese) and the WSJ corpus \cite{wsj} (in English). JNAS comprises roughly 80 hours from 320 speakers, and WSJ has about 70 hours of speech from 280 speakers.

In the multilingual training, the encoder and the attention module are shared among the Ainu ASR model and the models for other languages, and they are trained using data for all languages.
Figure 2 shows the architecture for the multilingual learning with two corpora. 
When the input acoustic features are from the Ainu ASR corpus, they go through the shared encoder and attention module and are delivered into the decoder on the left side in Figure 2 as a context vector. 
In this case, the right-side decoder is not trained.

\begin{figure}[t]
\centering
\includegraphics[width=1.0\columnwidth]{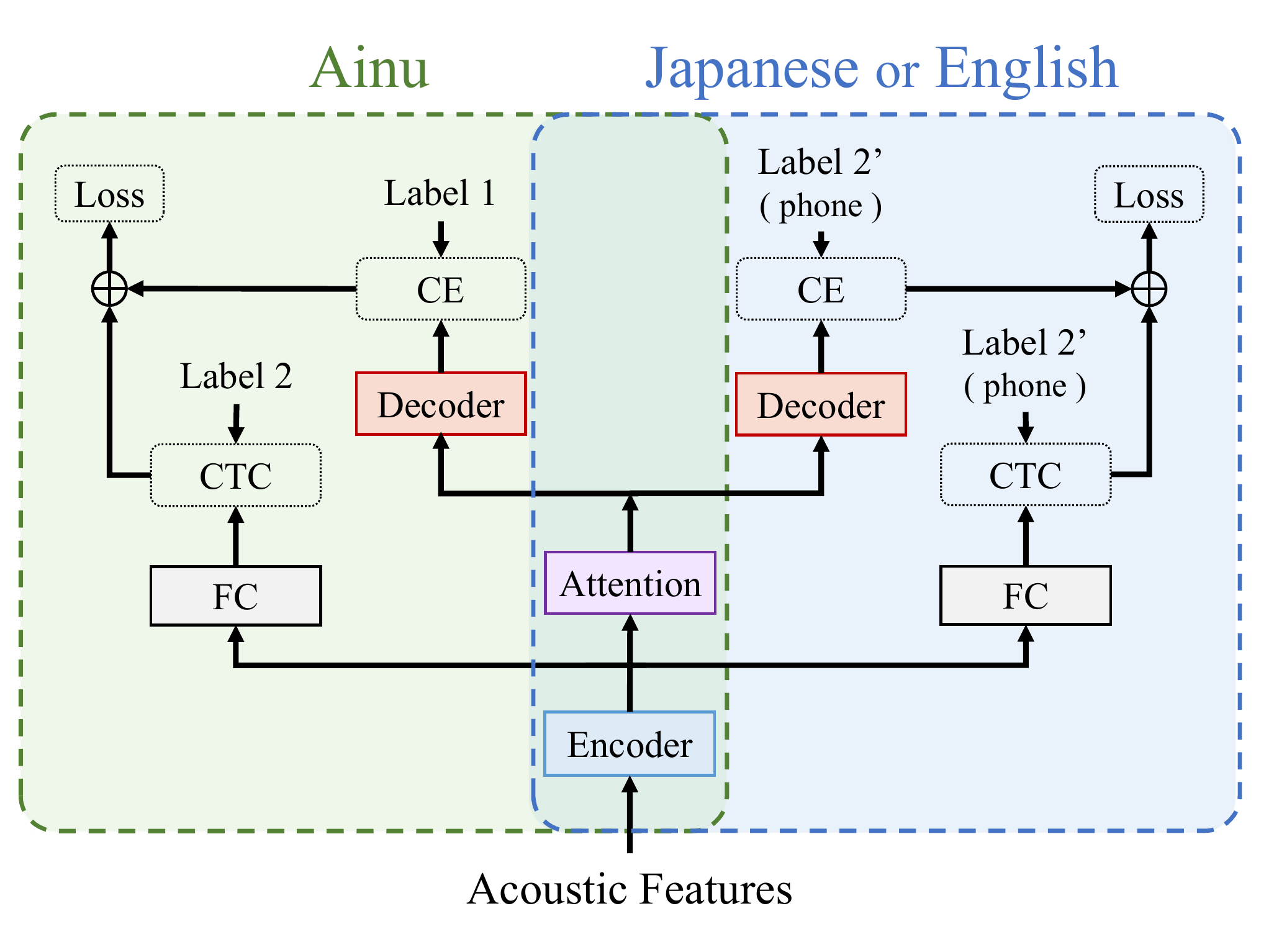}
\caption{The architecture of the multilingual learning with two corpora. `FC' and `CE' means `fully connected' and `cross-entropy' respectively.}
\label{fig:ctclas}
\end{figure}

\begin{table*}[t]
\centering
\caption{ASR performance for each speaker and modeling unit. The lowest error rates for each unit are highlighted.\vspace{5pt}}
\begingroup
\setlength{\tabcolsep}{7.7pt}
\renewcommand{\arraystretch}{1.15} 
\vspace{5pt}
\scalebox{1.0}[1.0]{
\begin{tabular}{c|c|c|rrrrrrrr:r}\hline\hline
 &  & \multicolumn{1}{c|}{units} & \multicolumn{1}{c}{KM}& \multicolumn{1}{c}{UT}& \multicolumn{1}{c}{KT}& \multicolumn{1}{c}{HS}& \multicolumn{1}{c}{NN}& \multicolumn{1}{c}{KS}& \multicolumn{1}{c}{HY}& \multicolumn{1}{c:}{KK}& \multicolumn{1}{c}{average}  \\ \hline
\multirow{8}{*}{speaker-closed}  & \multirow{4}{*}{WER (\%)} & phone   & 22.2& 28.5& 24.2& 28.6& 27.2& 30.6& 40.4& 36.1& 27.9\\
 &  & syllable   & \textbf{13.2} & \textbf{18.4} & \textbf{19.6} & 29.4& 26.7& 26.7& 38.9& \textbf{29.0} & \textbf{21.7} \\
 &  & WP   & 14.4& 20.0& 21.6& \textbf{25.0} & \textbf{27.1} & \textbf{23.2} & \textbf{37.8} & 42.5& 22.3\\
 &  & word   & 14.7& 19.6& 21.3& 32.9& \textbf{27.1} & 24.6& 40.7& 31.2& 23.1\\ \cline{2-12} 
 & \multirow{4}{*}{PER (\%)} & phone   & 10.7& 16.3& 7.9 & \textbf{5.6}  & \textbf{7.4}  & 13.6& 10.1& 14.8& 11.1\\
 &  & syllable   & \textbf{3.2}  & \textbf{6.9}  & \textbf{4.4}  & 7.7 & 7.9 & 9.5 & \textbf{9.4}  & \textbf{10.7} & \textbf{6.3}  \\
& & WP   & 4.7 & 8.0 & 5.2 & 6.7 & 8.4 & \textbf{6.8}  & 10.4& 12.6& 7.1 \\
 &  & word   & 11.2& 12.9& 12.6& 24.0& 17.1& 15.4& 27.0& 20.1& 15.9\\ \hline
\multirow{8}{*}{speaker-open} & \multirow{4}{*}{WER (\%)} & phone   & - & - & 38.8 & 40.5 & 41.9 & 53.1 & 35.9 & 54.7 & 43.4    \\
 &  & syllable   & - & - & \bf{33.4}& 37.8 & \bf{37.3} & \bf{47.2} & 32.0 & \bf{48.6}& \bf{38.6}    \\
 &  & WP   & - & - & 58.4 & \bf{37.2} & 38.6 & 47.9 & 32.6 & 48.8 & 45.7    \\
 &  & word   & -   & - & 34.0 & 49.0 & 39.4 & 48.9 & \bf{31.5} & 84.3 & 46.6    \\ \cline{2-12} 
 & \multirow{4}{*}{PER (\%)} & phone   & - & - & 14.9 & 13.9 & 15.9 & 21.4 & 11.2 & 27.0 & 17.1     \\
 &  & syllable   & - & - & \bf{10.7} & \bf{12.6} & \bf{13.5} & \bf{16.5} & \bf{10.3} & \bf{22.0}  & \bf{13.8}     \\
 &  & WP  & -& - & 41.5 & 14.1 & 15.9 & 19.3 & 11.5 & 23.6   & 23.6       \\
 &  & word   & - & - & 24.6 & 39.9 & 29.6 & 33.1 & 20.4 & 67.0   & 34.8     \\ \hline
\end{tabular}}
\endgroup
\end{table*}

\section{Experimental Evaluation}
In this section the setting and results of ASR experiments are described and the results are discussed.

\subsection{Data Setup}
The ASR experiments were performed in speaker-open condition as well as speaker-closed condition.

In the speaker-closed condition, two episodes were set aside from each speaker as development and test sets. Thereafter, the total sizes of the development and test sets turns out to be 1585 IPUs spanning 2 hours 23 minutes and 1841 IPUs spanning 2 hours and 48 minutes respectively.  The ASR model is trained with the rest data.
In the speaker-open condition, all the data except for the test speaker's were used for training 
As it would be difficult to train the model if all of the data of speaker KM or UT were removed, experiments using their speaker-open conditions were not conducted.

\subsection{Experimental Setting}
The input acoustic features were 120-dimensional vectors made by frame stacking \cite{DBLP:journals/corr/TianZMHW17a} three 40-dimensional log-mel filter banks features at contiguous time frames. The window length and the frame shift were set to be 25ms and 10ms. The encoder was composed of five BiLSTM layers and the attention-based decoder had a single layer of LSTM. Each LSTM had 320 cells and their weights were randomly initialized using a uniform distribution \newcite{DBLP:journals/corr/HeZR015} with biases of zero. The fully connected layers were initialized following $\mathcal{U}{(-0.1, 0.1)}$. The weight decay \cite{Krogh92asimple} whose rate was $10^{-5}$ and the dropout \cite{dropout} following $\mathcal{B}e(0.2)$ were used to alleviate overfitting. The parameters were optimized with Adam \cite{Adam}. The learning rate was $10^{-3}$ at first and was multiplied by $10^{-1}$ at the beginning of 31st and 36th epoch \cite{you2019does}. The mini-batch size was 30 and the utterances (IPUs) were sorted in an ascending order of length. To stabilize the training, we removed utterances longer than 12 seconds.

The loss function of the model was a linear sum of the loss from CTC and the attention-based decoder,
\begin{align}
\mathcal{L}_{\rm{all}} = \lambda \mathcal{L}_{\rm{attn}} + (1 - \lambda) \mathcal{L}_{\rm{CTC}},
\end{align}
\vspace{1pt}
where $\lambda$ was set to be 0.5. Through all experiments, the phone labels are used to train the auxiliary CTC task because it is reported that the hierarchical architecture, using few and general labels in the auxiliary task, improves the performance \cite{HMTL}. 

Strictly speaking, the number of each modeling units depends on the training set, but there are roughly 25-phone, 500-syllable, and 5,000-word units including special symbols that represent the start and end of a sentence. The words occurring less than twice were replaced with `$\langle$unk$\rangle$'. The vocabulary size for word piece modeling was set to be 500. These settings were based on the results of preliminary experiments with the development set.

For the multilingual training, we made three training scripts by concatenating the script of Ainu and other languages (JNAS, WSJ, JNAS and WSJ). The model was trained by these scripts until 30th epoch. From 31$^{\rm{st}}$ and 40th epoch, the model was fine-turned by the Ainu script.
Phone units are used for JNAS and WSJ throughout the experiments.

\subsection{Results}
Table 4 shows the phone error rates (PERs) and word error rates (WERs) for the speaker-closed and speaker-open settings. The `average' is weighted by the numbers of tokens in the ground truth transcriptions for speaker-wise evaluation sets.

The word recognition accuracy reached about 80\% in the speaker-closed setting. In the speaker-open setting it was 60\% on average and varied greatly from speaker to speaker (from 50\% to 70\%). The best phone accuracies in the speaker-closed and speaker-open settings were about 94\% and 86\%. Regardless of the settings, the syllable-based modeling yielded the best WER and PER. This suggests that syllables provide reasonable coverage and constraints for the Ainu language in a corpus of this size.

The PERs of the word unit model were larger than those of other units. This is because the word model often outputs the `$\langle$unk$\rangle$' symbols while other unit models are able to output symbols similar in sound as below.

\begin{table}[H]
\centering
\begingroup
\renewcommand{\arraystretch}{1.2} 
\begin{tabular}{c|l} 
\it{ground-truth} & i okake un a unuhu a onaha \\ 
\it{syllable model} & piokake un a unuhu a onaha \\
\it{word model} & $\langle$unk$\rangle$ un a unuhu a onaha \\ 
\end{tabular}
\endgroup
\vspace{-5pt}
\end{table}

\begin{table}[t]
\centering
\caption{Results of multilingual training.\vspace{5pt}}
\begingroup
\setlength{\tabcolsep}{12pt}
\renewcommand{\arraystretch}{1.15} 
\vspace{5pt}
\scalebox{1.0}[1.0]{
\begin{tabular}{c|l|r|r} \hline \hline
                      \multicolumn{2}{c|}{speaker-}             &\multicolumn{1}{c|}{closed}                       & \multicolumn{1}{c}{open}   \\ \hline
\multirow{4}{*}{WER (\%)} & Ainu  & 21.7   & 38.6          \\
                      & +~JNAS & \bf{21.1}   & 34.8              \\ 
 & +~WSJ & 21.3  &35.8         \\
                      & +~both   & 21.4 &\bf{34.7}   \\ \hline
                      \multirow{4}{*}{PER (\%)} & Ainu &6.3  &13.8             \\
                      & +~JNAS   & \bf{6.0} & 11.7    \\ 
 & +~WSJ & \bf{6.0} & 12.1            \\
                      & +~both   & \bf{6.0} & \bf{11.2}     \\ \hline
\end{tabular}}
\endgroup
\end{table}

In this example, the PER of the syllable model is 5\% and that of the word model is 30\% even though the WERs are the same. (The output of the syllable model is rewritten into words using the `$\langle$wb$\rangle$' symbol.)

WERs are generally much larger than PERs and it is further aggravated with the Ainu language. This is because, as mentioned in Section 2.1, the Ainu language has a lot of compound words and the model may be confused about whether the output is multiple words or a single compound word. The actual outputs frequently contain errors as below. The WER of this example is 57\% though the PER is zero.

\begin{table}[H]
\centering
\begingroup
\renewcommand{\arraystretch}{1.2} 
\begin{tabular}{c|l} 
\it{ground-truth} & nen poka apkas an mak an kusu \\ 
\it{output} & nenpoka apkas an makan kusu \\ 
\end{tabular}
\endgroup
\end{table}

The results of multilingual training in which the modeling unit is syllables are presented in Table 5. All error rates are the weighted averages of all evaluated speakers.  Here, `+~both' represents the result of training with both JNAS and WSJ corpora. The multilingual training is effective in the speaker-open setting, providing a relative WER improvement of 10\%. The JNAS corpus was more helpful than the WSJ corpus 
because of the similarities between Ainu and Japanese language.

\section{Summary}
In this study, we first developed a speech corpus for Ainu ASR and then, using the end-to-end model with CTC and the attention mechanism, compared four modeling units: phones, syllables, word pieces, and words. The best performance was obtained with the syllable unit, with which WERs in the speaker-closed and speaker-open settings were respectively about 20\% and 40\% while PERs were about 6\% and 14\%. Multilingual training using the JNAS improved the performance in the speaker-open setting. Future tasks include reducing the between-speaker performance differences by using speaker adaptation techniques.

\section{Acknowledgement}
The data sets used in this study are provided by the Ainu Museum and Nibutani Ainu Culture Museum. The authors would like to thank Prof. Osami Okuda of Sapporo Gakuin University for his useful advices on the Ainu language.

\section{References}
\label{main:ref}

\bibliographystyle{lrec}
\bibliography{my}

\end{document}